# A New IRIS Normalization Process For Recognition System With Cryptographic Techniques

Nithyanandam.S [1], Gayathri.K.S [2], Priyadarshini P.L.K [3]

[1,2,3] CSE Department, PRIST University,
Thanjavur, Tamilnadu, India

### Abstract

Biometric technologies are the foundation of personal identification systems. It provides an identification based on a unique feature possessed by the individual. This paper provides a walkthrough for image acquisition, segmentation, normalization, feature extraction and matching based on the Human Iris imaging. A Canny Edge Detection scheme and a Circular Hough Transform, is used to detect the iris boundaries in the eye's digital image. The extracted IRIS region was normalized by using Image Registration technique. A phase correlation base method is used for this iris image registration purpose. The features of the iris region is encoded by convolving the normalized iris region with 2D Gabor filter. Hamming distance measurement is used to compare the quantized vectors and authenticate the users. To improve the security, Reed-Solomon technique is employed directly to encrypt and decrypt the data. Experimental results show that our system is quite effective and provides encouraging performance.

## 1. The IRIS

The human iris recently has attracted the attention of biometrics-based identification and verification research and development community. The iris is so unique that no two irises are alike, even among identical twins or even between the left and right eye of the same person, in the entire human population.

The term iris recognition[1] refers to identifying an iris image by computational algorithms and it is used as an identity. Iris recognition technology offers the highest accuracy in identifying individuals as compared to any other method available. A key advantage of iris recognition is its stability, or template longevity, as, barring trauma, a single enrolment can last a lifetime.

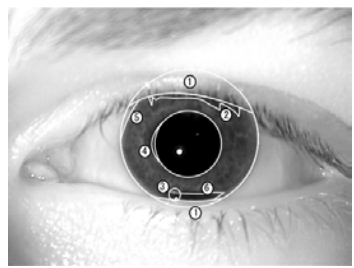

Fig 1. Image of IRIS. The numbers in this Fig correspond to, 1) eyelids, 2) eyelashes, 3) specular reflections, 4) pupil, 5) sclera, and 6) shadow caused by an eyelid.

The human different characteristic can be used as a biometric characteristic as long as it satisfies the following requirements:

1. Universality    2. Distinctiveness    3. Permanence
4. Collectability  5. Performance        6. Acceptability
7. Circumvention

Comparison of some of the biometric identifiers based on seven factors is provided in Fig. 2.

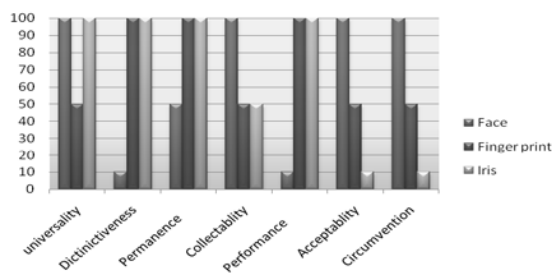

Fig 2. Comparison of some of the biometric identifiers

## 2. IRIS Recognition

Image pre-processing and normalization is significant part of iris recognition systems. The stages involved in Iris reorganization are,

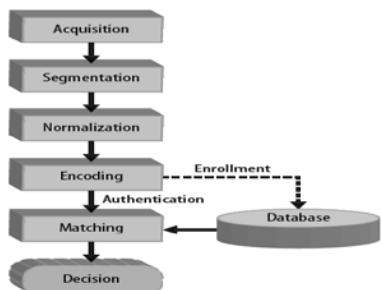

Fig 3. Frame work of iris recognition

As Fig 3 shows, most iris recognition systems consist of five basic modules leading to a decision:

- The *acquisition* module obtains an image of the eye.
- The *segmentation* module localizes the iris's spatial extent in the eye image by isolating it f rom other structures in its vicinity, such as the sclera, pupil, eyelids, and eyelashes.
- The *normalization* module invokes a g eometric normalization scheme to transform the segmented iris image from cartesian coordinates to polar coordinates.
- The *encoding* module uses a feature-extraction routine to produce a binary code.
- The *matching* module determines how closely the produced code matches the encoded features stored in the database.

## 3. Segmentation

The first stage of iris recognition is to isolate the actual iris region in a digital eye image. The iris region is approximated by two circles, one for the iris/sclera boundary and another for the iris/pupil boundary. The eyelids and eyelashes normally occlude the upper and lower parts of the iris region.

### 3.1. Canny Edge Detection and Localization

Canny edge detection is used to create an edge map[2]. The boundary of the iris is located by using canny edge detection technique. These parameters are the centre coordinates x and y, the radius r, which are able to define any circle according to the equation,

$$x^2 + y^2 = r^2 \qquad (1)$$

In performing the preceding edge detection step, the derivatives of the horizontal direction is to detect the eyelids, and the vertical direction is to detect the outer circular boundary of the iris.

### 3.2. Hough Transform

The Hough transform is a feature extraction technique used in image analysis.

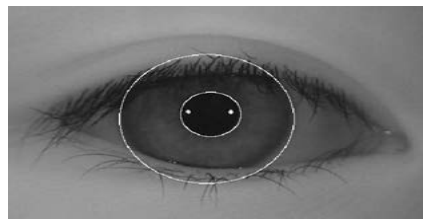

Fig 4. Detection of circular boundaries of pupil and iris.

The parabolic Hough transform is used to detect the eyelids, approximating the upper and lower eyelids with parabolic arcs, which are represented as,

$$(-(x-h_j)\sin\theta_j + (y-k_j)\cos\theta_j)^2 = (a_j(x-h_j)\cos\theta_j + (y-k_j)\sin\theta_j) \qquad (2)$$

Where, $a_j$ controls the curvature, $(h_j, k_j)$ is the peak of the parabola and $\theta_j$ is the angle of rotation relative to the x-axis.

## 4. Normalization

Once the segmentation module has estimated the iris's boundary, the normalization module uses image registration technique to transform the iris texture from cartesian to polar coordinates. The process, often called *iris unwrapping*, yields a rectangular entity that is used for subsequent processing.

*Normalization has three advantages:*

- It accounts for variations in pupil size due to changes in external illumination that might influence iris size.
- It ensures that the irises of different individuals are mapped onto a common image domain in spite of the variations in pupil size across subjects.
- It enables iris registration during the matching stage through a simple translation operation that can account for in-plane eye and head rotations.

Associated with each unwrapped iris is a binary mask that separates iris pixels (labeled with a "1") from pixels that correspond to the eyelids and eyelashes (labeled with a "0") identified during segmentation. After normalization, photometric transformations enhance the unwrapped iris's textural structure.

### 4.1 Image Registration Techniques

Image registration is important in remote sensing, medical imaging, and other applications where images must be aligned to enable quantitative analysis or qualitative comparison.

*Steps involved in Image Registration*

Image registration essentially consists of following steps as per Zitova and Flusser.

- *Feature detection:* Salient and distinctive objects in both reference and sensed images are detected.
- *Feature matching:* The correspondence between the features in the reference and sensed image established.
- *Transform model estimation:* The type and parameters of the so-called mapping functions, aligning the sensed image with the reference image, are estimated.
- *Image resampling and t ransformation:* The sensed image is transformed by means of the mapping functions.

### 4.2 Mapping Functions

The system employs an image registration technique, which geometrically warps a newly acquired image, $I_a(x,y)$ into alignment with a s elected database image $I_d(x,y)$. When choosing a mapping function $(u(x,y),v(x,y))$ to transform the original coordinates, the image intensity values of the new image are made to be close to those of corresponding points in the reference image. The mapping function must be chosen so as to minimize,

$$\iint (I_d(x,y) - I_a(x-u,y-v))^2 \, dxdy \qquad (3)$$

while being constrained to capture a similarity transformation of image coordinates (x,y) to (x',y') that is,

$$\begin{bmatrix} x \\ y \end{bmatrix} = \begin{bmatrix} x' \\ y' \end{bmatrix} - s\,R(\Phi) \begin{bmatrix} x \\ y \end{bmatrix} \qquad (4)$$

with *s* as a scaling factor and *R(Φ)* is a matrix representing rotation by Φ. In implementation, given a pair of iris images $I_a$ and $I_d$, the warping parameters s and Φ are recovered via an iterative minimization procedure.

### 4.3 Phase Correlation Method

The idea behind this method is quite simple and is based on the Fourier shift property, which states that a shift in the coordinate frames of two functions is transformed in the Fourier domain as linear phase differences.

*Principle of phase correlation:*

Firstly, we briefly introduce the principle of phase correlation[4]. Suppose that there is a shift$(d_1,d_2)$ between two images $S_k$ and $S_{k+1}$,

$$S_k(n_1,n_2) = S_{k+1}(n_1+d_1, n_2+d_2) \qquad (5)$$

The shift in the spatial domain is thus reflected as a space change in the frequency domain. We can obtain the complex-valued cross-power spectrum expression,

$$C_{k,k+1}(f_1,f_2) = S_{k+1}(f_1,f_2)\, S_k^*(f_1,f_2) \qquad (6)$$

Where, * denotes the complex conjugate. To reduce the influence of luminance variation, the right side of (6) is normalized as follows,

$$\Phi C_{k,k+1}(f_1,f_2) = \frac{S_{k+1}(f_1,f_2)\, S_k^*(f_1,f_2)}{|S_{k+1}(f_1,f_2)\, S_k^*(f_1,f_2)|} \qquad (7)$$

From (6,7) we can obtain the following equation,

$$C_{k,k+1}(n_1,n_2) = \delta(n_1+d_1, n_2+d_2) \qquad (8)$$

Where, δ denotes the pulse function. The above equation indicates that if we confine the location pulse in the cross correlation map of two images, we could obtain the spatial displacement between. One example of the cross correlation map shown in Fig.5

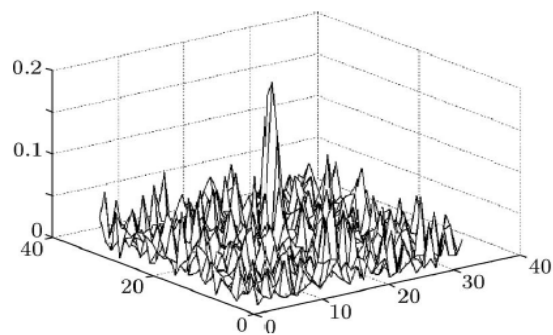

Fig.5 Phase correlation map

## 5. Encoding

Although a recognition system can use the unwrapped iris directly to compare two irises, most systems first use a feature extraction routine to encode the iris's textural content. In a commonly used encoding mechanism, 2D Gabor wavelets are first used to extract the local phasor information of the iris texture.

*Gabor Filters*

Gabor filters are able to provide optimum conjoint representation of a signal in space and spatial frequency[2]. A Gabor filter is constructed by modulating a sine/cosine wave with a Gaussian.

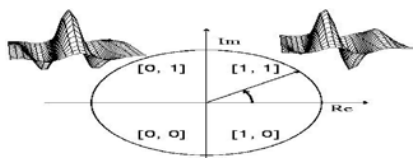

Fig.6 A quadrature pair of 2D Gabor filters

Daugman makes uses of a 2D version of Gabor filters in order to encode iris pattern data. A 2D Gabor filter over an image domain (x,y) is represented as,

$$G(x,y) = e^{(-\pi [[x-x_0]^2 \alpha^2 + (y-y_0)^2 \beta^2] - 2\pi i[u_0(x-x_0)+v_0(y-y_0)])} \quad (9)$$

where, $(x_0,y_0)$ specify position in the image, $(\alpha,\beta)$ specify the effective width and length, and $(u_0,v_0)$ specify modulation, which has spatial frequency $t(u_0^2 + v_0^2)$.

## 6. Matching

The matching module generates a match score by comparing the feature sets of two iris images. One technique for comparing two IrisCodes is to use the Hamming distance, which is the number of corresponding bits that differ between the two IrisCodes.

*Hamming Distance*

The Hamming distance gives a measure of how many bits are the same between two bit patterns. In comparing the bit patterns X and Y, the Hamming distance (HD)[6] is defined as the sum of disagreeing bits (sum of the exclusive-OR between X and Y) over N, the total number of bits in the bit pattern. The derived equation for HD is,

$$HD = \frac{1}{N} \sum_{j=1}^{N} X_i (XOR) Y_i \quad (10)$$

Since an individual iris region contains features with high degrees of freedom, each iris region will produce a bit-pattern which is independent to that produced by another iris. On the other hand, two iris codes produced from the same iris will be highly correlated.

If two patterns are derived from the same iris, the Hamming distance between them will be close to 0.0, since they are highly correlated and the bits should agree between the two iris codes.

## 7. Biometric Security Systems

7.1 Introduction

A biometric system is essentially a pattern recognition system that operates by acquiring biometric and providing more security by using an encryption and decryption techniques. Depending on the application context, a biometric system may operate either in verification mode or identification mode[3].

7.2 Encryption and Decryption

Encryption is the process of transforming information (referred to as plaintext) using an algorithm (called cipher) to make it unreadable to anyone except those possessing special knowledge, usually referred to as a key[5]. The result of the process is encrypted information (in cryptography, referred to as ciphertext).

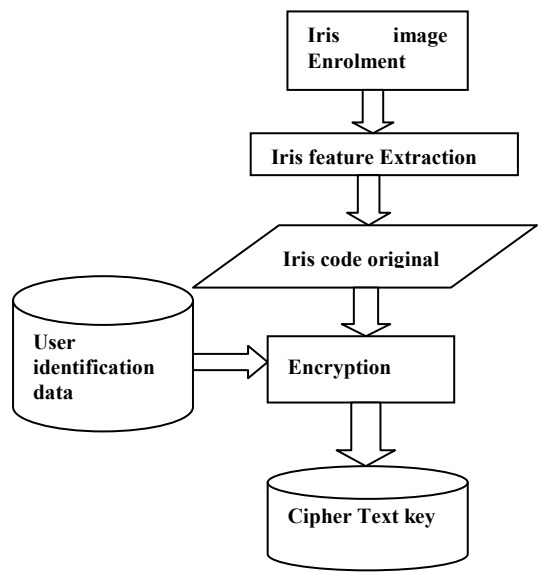

Fig.7 Encryption Technique

Decryption is the process of decoding data that has been encrypted into a secret format[7]. Decryption requires a secret key or password. There are of course a wide range of cryptographic algorithms in use.

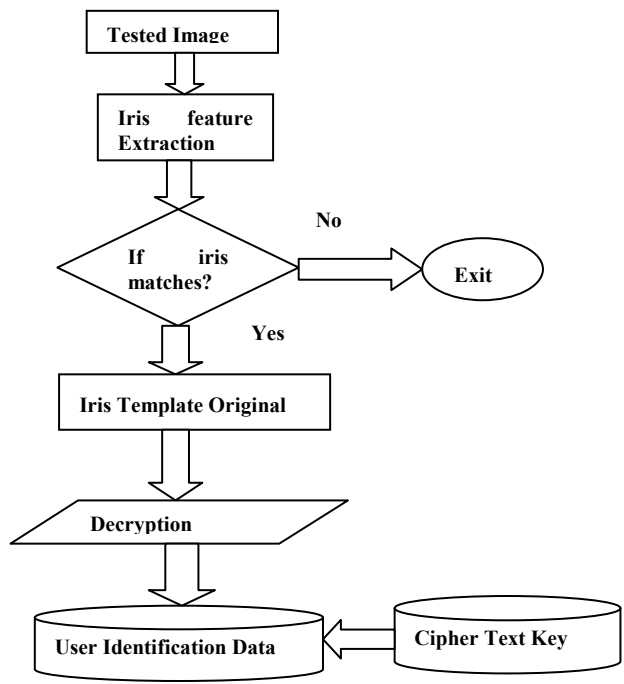

Fig. 8 Decryption Technique

### 7.3 Hadamard and Reed-Solomon Approach

Hadamard error correction works well for random errors. To do this, we create an orthogonal Hadamard code $H_c$ with $2_{k-1}$ columns and $2_k$ rows. The iris code is then split into groups of k bits – each group of k bits is then converted from binary to decimal and used to indicate a specific row of the Hadamard matrix. The -1s are then treated as 0s to allow for XOR. This new matrix, represented by various rows of the Hadamard matrix, is then used to XORed with the iris code.

When decoding, we find the dot product of each string of $2_{k-1}$ bits and every row of the Hadamard matrix. Since the Hadamard matrix is orthogonal, the row with the max value after the dot product should be the same row that was used to represent the code. Using this row, we can convert the decimal value back to the binary to recover the original key, k bits at a time. Since Hadamard error correction code wasn't strong enough to completely recalculate the key, additional Reed Solomon codes were used since it is robust against burst errors that occur with obstruction of the iris by eyelashes, eyelids, etc.

The original concept of Reed-Solomon coding describes encoding of k message symbols by viewing them as coefficients of a polynomial p(x) of maximum degree k-1 over a finite field of order N, and evaluating the polynomial at n>k distinct input points. Sampling a polynomial of degree k-1 at more than k points creates an over determined system, and allows recovery of the polynomial at the receiver given any k out of n s ample points using (Lagrange) interpolation. The sequence of distinct points is created by a generator of the finite field's multiplicative group, and includes 0, thus permitting any value of n up to N.

## 8. Experimental Results

### 8.1 Segmentation

*Iris Segmentation:* This involves first employing Canny Edge Detection to generate an edge map. In the work, in order to increase the overall speed of the system, circle houghman detection algorithm is used for *Iris Localization*. The output provided at the segmentation level is as follows,

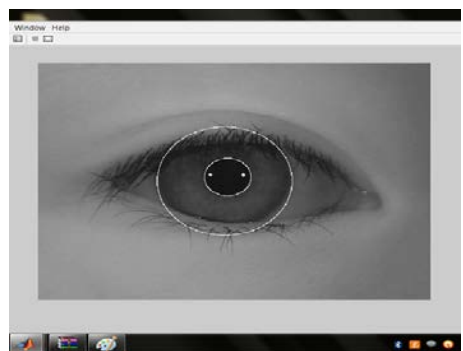

Fig 9. Detection of circular boundaries of pupil and iris.

### 8.2 Normalization

In our work phase correlation image registration method is used and following output is retrieved,

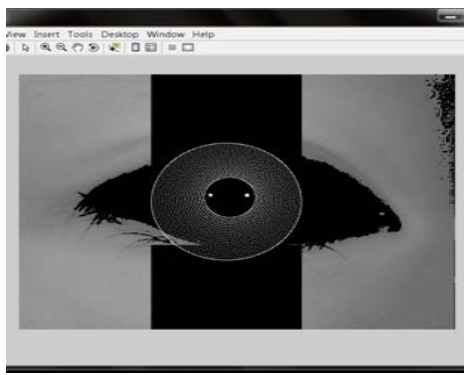

Fig 10. Normalized iris image

The noise present in iris can be detected as follows,

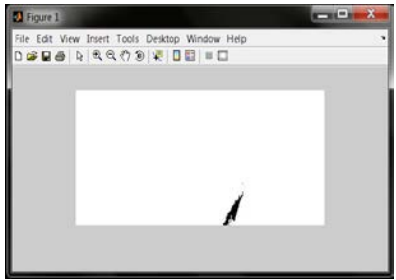

Fig.11 Polar noise detection

The efficient result provided by phase correlation method is high and it can be proved by following graph,

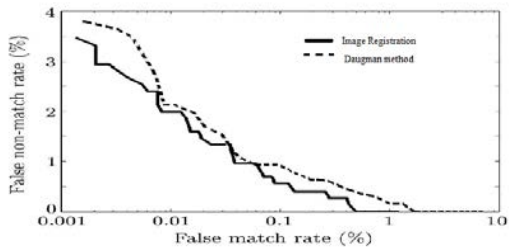

Fig.12 Verification Result

### 8.3 Feature Encoding

The encoding process produces a bitwise template containing a number of bits of information, and a corresponding noise mask which corresponds to corrupt areas within the iris pattern, and marks bits in the template as corrupt.

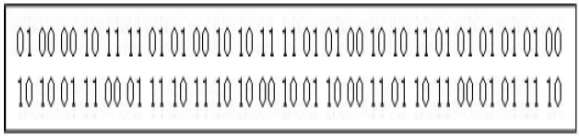

Fig.13 The sample template obtained in feature encoding

A histogram is a tool that allows you to visualize the proportion of numbers that fall within a given bin, or interval.

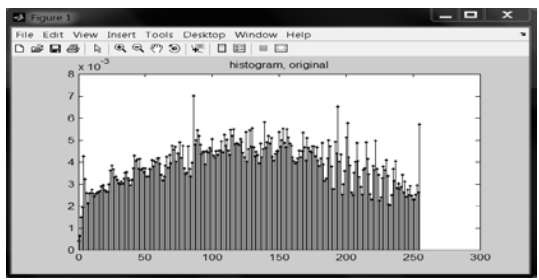

Fig.14 Histogram for an iris image

### 8.4 Matching

For matching, the Hamming distance was chosen as a metric for recognition, since bit-wise comparisons were necessary.

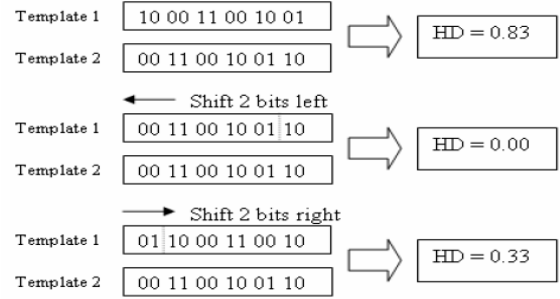

Fig.15 An illustration of the shifting process.

### 8.5 Security

It is found that a 60% accuracy rate for a single correct accept between two different pictures of the same iris could be achieved using our implementation of Hadamard and Reed Solomon error correction codes. If multiple images for a single iris were to be stored for matching and comparison, the percentage of correct acceptance could rise to 93.6% for correct acceptance.

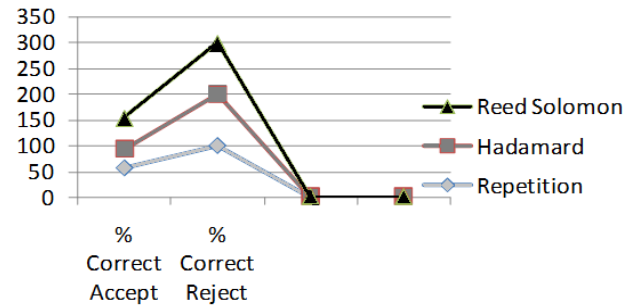

Fig.16 Comparison of different error correction techniques

## 9. Conclusions

Registration is a method used to determine the geometrical transformation that aligns points in one view of an object with corresponding points in another view of that object or another object. We have successfully developed a new Iris recognition system using phase correlation technique and further comparison of two digital eye-images has been done. These techniques have been independently studied for several different applications, resulting in a large body of research. The security is increased by using encryption techniques. Experimental results show that the proposed algorithm has an encouraging performance.

**Nithyanandam.S** is an Asst. Professor in the CSE Department at PRIST University. He received the M.Tech Degree in Information from Punjabi University Technology, Patiala in 2003, MCA., M.Sc(Phy.), M.Phil. Degree from Bharathidasan University, Tiruchirappalli. His Current area of interest are Image Recognition, Biometric Authentication and OFDM. He has Presented 6 Papers in international / National Conferences and also published 3 Paper in International Journals. He also published a book entitled Computer Networks, Published by SAMS Publishers.

**Gayathri.K.S** is a Asst. Professor in the CSE Department at PRIST University. She received the M.E Degree in Computer Science Engineering from Anna University, Chennai in 2009 and B.E (ECE) Degree from Anna University, Chennai. She has Presented 5 Papers in international / National Conferences and also published one Paper in International Journals.

**Priyadarsini. P.L.K** is a Professor in the CSE Department at PRIST University. She has obt ained her Ph.D Degree in Computer Science from NIT, Tiruchirappalli in the year 2009 and MCA Degree from Sri Padmavathi Mahila Viswa Vidyalayam, Tirupathi. She has Presented 6 Papers in International / National Conferences and also published five Papers in International Journals.